\documentclass[letterpaper, 10 pt, conference]{ieeeconf}  

\IEEEoverridecommandlockouts                              

\usepackage{graphics} 
\usepackage{epsfig} 
\usepackage{mathptmx} 
\usepackage{times} 
\usepackage{amsmath} 
\usepackage{amssymb}  
\usepackage{multirow}
\usepackage{breqn}
\usepackage{algorithm}
\usepackage{algpseudocode}
\usepackage{caption}
\usepackage{booktabs}
\usepackage{hyperref}
\usepackage{float}

\title{\LARGE \bf
Bayesian Intention for Enhanced Human Robot Collaboration}

\author{Vanessa Hernandez-Cruz, Xiaotong Zhang, and Kamal Youcef-Toumi
\thanks{All authors are with the Mechatronics Research Laboratory, Massachusetts Institute of Technology, Cambridge, MA, 02139, USA
{\tt\small \{vanessa7, kevxt, youcef\}@mit.edu }}
\thanks{This research was made possible by the support and partnership
of King Abudlaziz City for Science and Technology
(KACST) through the Center for Complex Engineering
Systems at Massachusetts Institute of Technology (MIT) and
KACST.}
}
\begin{document}
\maketitle
\thispagestyle{empty}
\pagestyle{empty}

\begin{abstract}
Predicting human intent is challenging yet essential to achieving seamless Human-Robot Collaboration (HRC). 
Many existing approaches fail to fully exploit the inherent relationships between objects, tasks, and the human model. Current methods for predicting human intent, such as Gaussian Mixture Models (GMMs) and Conditional Random Fields (CRFs), often lack interpretability due to their failure to account for causal relationships between variables. To address these challenges, in this paper, we developed a novel Bayesian Intention (BI) framework to predict human intent within a multi-modality information framework in HRC scenarios. This framework captures the complexity of intent prediction by modeling the correlations between human behavior conventions and scene data. Our framework leverages these inferred intent predictions to optimize the robot's response in real-time, enabling smoother and more intuitive collaboration. We demonstrate the effectiveness of our approach through a HRC task involving a UR5 robot, highlighting BI's capability for real-time human intent prediction and collision avoidance using a unique dataset we created. Our evaluations show that the multi-modality BI model predicts human intent within 2.69ms, with a 36\% increase in precision, a 60\% increase in F1 Score, and an 85\% increase in accuracy compared to its best baseline method. The results underscore BI's potential to advance real-time human intent prediction and collision avoidance, making a significant contribution to the field of HRC.
\end{abstract}


\section{Introduction}

Human intention prediction is crucial for safe and efficient Human-Robot Collaboration (HRC), especially in applications like collaborative assembly \cite{malik2019complexity}, smart warehouses \cite{9981064}, smart cities \cite{9761889}, and education \cite{xia2021modular}. By predicting human intentions, robots can dynamically update their behavior in real-time for smoother, safer, and more proactive HRC \cite{kothari2023enhanced}.

Moreover, human intention prediction plays a pivotal role in a new concept developed by our lab, called relevance, requiring human intention and objective as essential inputs \cite{zhang2024relevancehumanrobotcollaboration} \cite{zhang2024does}. The new task relevance is to determine the relevant sets of classes or elements within a scene based on their contributions to the human's tasks or objective in an HRC setup \cite{youtube_video}. Relevance allows the robot to discern which parts of a scene are essential for its immediate goals, which are supplementary, and which can be disregarded, therefore optimizing computational resources and enhancing interaction efficiency and quality. 
The human intention prediction framework developed in this paper can be applied to relevance quantification and, thus, has a broader significance. 

\begin{figure*}[t!]
\centering
\includegraphics[width= 0.95\textwidth]
{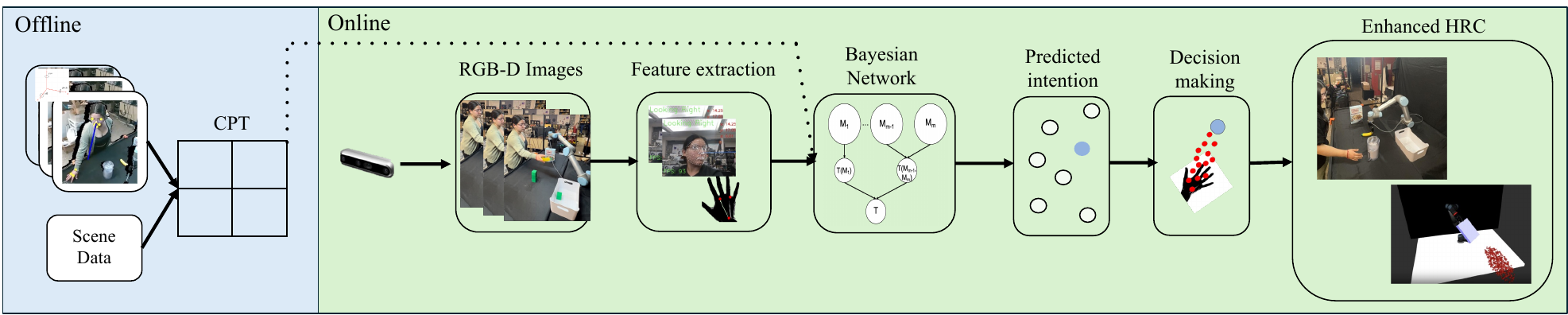}
\caption{The framework of Bayesian Intention and its application for enhanced HRC. The offline portion of the framework feeds a learned conditional probability table (CPT) to the online portion that predicts human intent by modeling causal relationships between the human and the scene where the robot changes its path and task plan accordingly.}
\label{fig:impl}
\end{figure*}

For efficient and accurate human intention prediction, we developed a novel framework that synthesizes information based on a comprehensive top-down and bottom-up approach through a Bayesian Network (BN). A top-down approach uses prior knowledge and expectations to interpret information. A bottom-up approach is based on current data without prior relationships in mind. The BN represents the top-down aspect. The heuristics represent the bottom-up approach with the streams of data that fuse into the BN nodes to update probabilities. In our BN, shown in Fig. \ref{fig:impl}, the parent nodes are derived from human conventions. Our BN uses m distinct modalities of information, $\mathrm{M}_{1}, \ldots, \mathrm{M}_{m}$. The nodes T($\mathrm{M}_{i}$) corresponds to the target based only on the i-th modality. The target, T, corresponds to the human's main object of interest. In this paper's HRC demonstration, these modalities are head orientation, hand orientation, and hand velocity vectors relative to the objects in the scene.

To the best of our knowledge, our methodology is unique in human intention prediction. First, we utilized the head orientation as a modality of information instead of gaze in other papers. Compared with gaze, head orientation can be extracted far more robustly and reliably from a distance and from diverse angles. Second, many human intent prediction works make a one-to-one correlation between intention and a single modality of information. To compensate for the discrepancy between head orientation and true human intention, we proposed an information fusion methodology based on BN with information of head orientation, hand orientation based on object affordance, and hand movement. 
In our work, we show that relying only on one modality of information is not enough to accurately and reliably predict human intent.

 We demonstrate and validate our method on a tabletop pick and place scenario of making a bowl of cereal with both real-life demonstration and simulation with a novel dataset collected. In the demonstration, we developed a method to apply the predicted intention to construct a virtual obstacle for dynamic robot task adaption and proactive motion planning. Enabling robots to foresee where a human is going based on intent derived from a BN, in real-time, significantly enhances HRC and reduces latency in completing tasks. 

To summarize, the contributions of this paper are as follows: (1) We developed and proposed a novel methodology and framework for human intent prediction based on multi-modality indicators derived from human behavior conventions; (2) We applied our real-time intent prediction methodology to create virtual obstacles with a proactive motion planner for human-robot collaboration; (3) We developed and collected a novel dataset that contains both the object information, including grasping affordance and world coordinate location, along with the human head and hand keypoints; (4) We demonstrate and validate our methodologies through simulation and real-world experiments. The results show the effectiveness of our intent prediction framework and its applicability for human-robot collaboration even with unforeseen or inattentive movements. 

\section{Related Works}

Human intention prediction in HRC has been studied using various approaches and information modalities.This section reviews existing approaches, their limitations, and how our work addresses these gaps.

\subsection{Information Modality for Intention Prediction}
Gaze is a popular modality of information for intent prediction because gaze direction provides direct and immediate cues about where a person is focusing on \cite{gazebasedintent,assistiverobo}. However, head orientation is more advantageous because it can be robustly and reliably detected with a camera from a further distance and from more angles since a head occupies significantly more pixels than just eyes. The head orientation can even be detected with a camera behind the person. In addition, head orientation allows for a richer contextual understanding of the social dynamics at play. For example, there may be a scenario where a person's head is tilted because they are focusing on a task, yet their visual eye direction may be directed towards a person they are interacting with. In our work, we reinforce human intent prediction by formalizing the connections between head orientation, hand orientation from object affordances, and hand movement. 

\subsection{Probabalistic Models for Intent Prediction}
Probabilistic models have been widely used for human intention prediction. GMMs identify clusters from data to generate a mixed Gaussian distribution, but they do not handle temporal dependencies between variables, making them susceptible to incorrect relationships \cite{shahpaper}. Additionally, GMMs assume that data clusters follow a normal distribution, which may not always be the case in HRC \cite{mainprice}. While methods like \cite{gmmpaper} use unsupervised learning to improve clustering, GMMs lack directionality and assume each data point independently comes from a Gaussian distribution. In contrast, our BI framework allows our BN to make more interpretable intent predictions by incorporating temporal dependencies and by capturing the causal relationships between variables. 

CRFs are undirected models that attempt to represent relationships between variables without considering causality. For instance, Koppula et al. \cite{crfp1} predict action labels using CRFs based on human activity and object affordances. However, because CRFs lack directionality, they fail to model causal relationships, which are crucial for understanding the influence between variables in real-time tasks. Approaches like \cite{crfp2} improve CRF models by reducing supervised training requirements, but they still fall short in providing interpretable, causal predictions. Our BN approach, on the other hand, explicitly models these causal connections, enhancing intent prediction reliability in dynamic HRC scenarios.

\section{Problem Definition and Methodology}
In our problem, a set of objects $\mathbf{O} = \{o_1, o_2, \ldots ,o_n\}$ is identified with an object detection algorithm, where $n$ represents the number of objects in the scene. The human intent encompasses one object in the scene denoted as $T$. Once $T$ is predicted from the three streams of information in the BN of Fig. \ref{fig:impl}, a virtual obstacle is constructed between the working hand and the intended target for the decision making algorithms to generate and update the plan and/or trajectories dynamically. Our methodology allows the robot to infer the short-term goal of the human's movement and react proactively.

\subsection{Modalities of Information}
Our BI framework uses three modes of information to model and quantify intent:


\subsubsection{Head Orientation}
\begin{figure}[t]
\centering
\includegraphics[width= \columnwidth]
{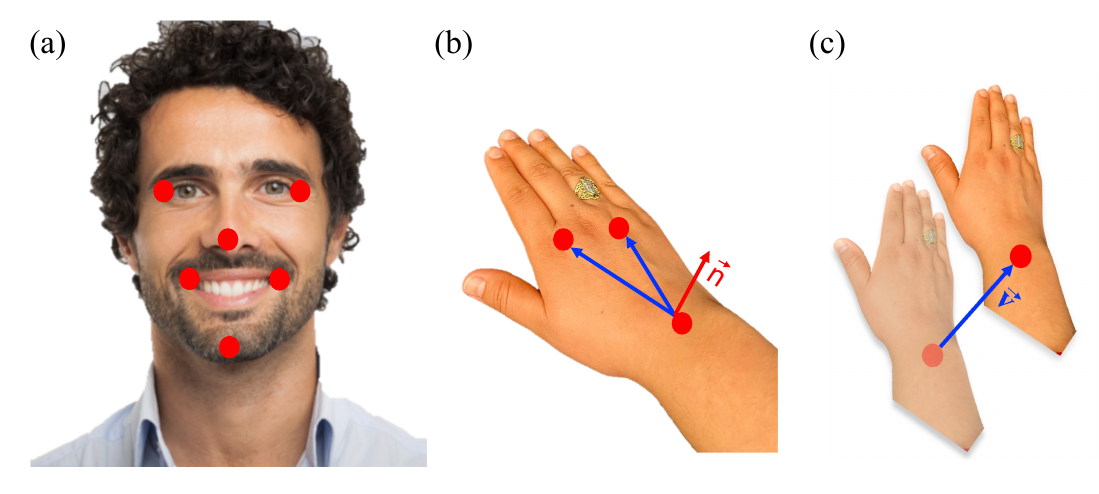}
\caption{The key points used to derive the features of the three modalities of information for (a) head orientation; (b) hand orientation; (c) hand motion. The head shot is from \cite{mediapipe}.}
\label{fig:handorien}
\end{figure}

Head orientation is an important feature for human intent prediction because it often correlates with the direction of a person's gaze,  a strong indicator of a person's focus. Unlike gaze, head orientation can be extracted from a long distance or from more angles, increasing the robustness and applicability of the algorithm. The head orientation vector ${\mathbf{h}}$ is calculated by using the six key points on the face shown in Fig. \ref{fig:handorien}(a). The points are deprojected from the camera frame into the world frame. An OpenCV algorithm \cite{headovideo} extracts the head orientation as a vector from the head's rotation angles. 

We define $\theta_{o_{i}}$ as the angle between the vector connecting the human's nose to the object $o_i$ and the head orientation vector. $\theta_{o_{i}}$ can be calculated as follows:

\begin{equation}
\theta_{o_i} = \arccos\left(\frac{{\mathbf{h}} \cdot {\mathbf{o}_{h_i}}}
{\left\|{\mathbf{h}}\right\| \cdot \left\|{\mathbf{o}_{h_i}}\right\|}\right)
\label{alg:dotprod}
\end{equation}

where ${\mathbf{o}}_{h_i}$ is the vector from the tip of the human's nose to $o_i$.

\subsubsection{Hand Orientation}

Humans adjust their hand orientation based on an object's visual affordance \cite{benchmarkintention}. Hand orientation is a valuable indicator that highlights a set of objects whose shape is conducive to the given hand orientation.

To estimate the hand orientation, we track three points on the hand as shown in Fig. \ref{fig:handorien}(b). We form two vectors from the three points that allow their cross-product to estimate the hand's orientation. This informs us whether the hand has a flexion orientation, which is used for objects that can be grasped comfortably with a relatively horizontal palm, or if the hand is oriented with a neutral orientation favorable towards vertical grabs. We define $\gamma$ as the angle between the normal vector of the user's hand ${\mathbf{n}}$ and the unit vector ${\mathbf{z}}$ in the world coordinate, which can be calculated as:


\begin{equation}
\gamma = \arccos\left(\frac{{\mathbf{n}} \cdot {\mathbf{z}}}
{\left\|{\mathbf{n}}\right\| \cdot \left\|{\mathbf{z}}\right\|}\right)
\end{equation}

Once $\gamma$ passes $\gamma_{h}$, the threshold angle for a hovering flexion position, the hand changes to a neutral state. Similarly, once $\gamma$ passes $\gamma_{v}$, the neutral threshold angle, the hand orientation is estimated as other, which means that the hand orientation is neither in a flexion nor in a neutral orientation. 

\subsubsection{Hand Motion}

The fulfillment of this intention is realized through the motion of the hand. Thus, understanding hand motion is crucial for accurately predicting human intent. The hand velocity ${\mathbf{v}}$ is estimated with the hand locations at two consecutive time steps. 
We define $\beta_{o_i}$ as the angle between the hand velocity and ${\mathbf{o}_{v_i}}$, the vector from the working wrist to the object $o_i$. $\beta_{o_i}$ can be calculated as:

\begin{equation}
\beta_{o_i} = \arccos\left(\frac{{\mathbf{v}} \cdot {\mathbf{o}_{v_i}}}
{\left\|{\mathbf{v}}\right\| \cdot \left\|{\mathbf{o}_{v_i}}\right\|}\right)
\label{eqn:betaeq}
\end{equation}


With the above methods, the features from the three modalities of information are computed and extracted efficiently.

\subsection{Target Prediction}

The features from the three modalities of information are integrated into a vector ${\mathbf{E}_t}$ for each time step: 

\begin{equation}
{\mathbf{E}_{t}} =[
\theta_{o_1}, ... \ \theta_{o_n}, \ 
\beta_{o_1}, ... \ \beta_{o_n}, \  
\gamma]^\top
\label{eqn:experiences}
\end{equation}

where $\top$ is the transpose operation. The dimension of ${\mathbf{E}_{t}}$ depends on the number of features available from the scene objects for each modality. The three modes of information we chose are based on a tabletop HRC scenario. However, the selection of features in ${\mathbf{E}_t}$ can be generalized to other forms of information available from any HRC scenario. Our Bayesian model utilizes ${\mathbf{E}_{t}}$ to aggregate the features available.

We formulate the human intent prediction in a probabilistic manner and solve it with a BN. In our formulation, the posterior probability, which is used for the final prediction, is derived as follows:

\begin{equation}
P_t(T_{o_{i}}|{\mathbf{E}}_{t}) = \frac{P_t({\mathbf{E}}_{t} | T_{o_{i}}) P_t(T_{o_{i}})}{P_t({\mathbf{E}}_{t})}  \\
\label{eqn1}
\end{equation}

where $T_{o_i}$ represents the event that object $o_i$ is the human target, $P_t({\mathbf{E}}_{t})$ is the probability of the evidence, $P_t(T_{o_i})$ is the prior probability, and $P_t({\mathbf{E}}_{t}|T_{o_{i}})$ is the conditional probability (CPT) of the evidence given the human target is $o_i$, which is learned from our dataset as shown in Fig. \ref{fig:impl}.

The prior probability in \ref{eqn1} is dynamically updated as:

\begin{equation}
P_t(T_{o_{i}}) = \begin{cases}
                1/n & \text{if t = 0}  \\
                P_{t-1}(T_{o_{i}} | \mathbf{E}_{t-1}) & \text{if t \textgreater{} 0} 
                \end{cases}
\label{eqn:cases}
\end{equation}

which means the prior probability $P_t( T_{o_{i}})$ updates to the last time step's posterior probability distribution except for the first time step. 

The target \( T \) is predicted by finding the object \( o_i \) from the set of all objects \( O \) that maximizes the posterior probability given the evidence \( \mathbf{E}_{t} \), which is represented mathematically as follows:

\begin{equation}
T = \arg\max_{o_i \in O} P_t(T_{o_i} | \mathbf{E}_{t})
\end{equation}

This methodology enables the robot to make accurate real-time predictions about human intentions.


\subsection{Decision making with human intention}
With the predicted human intention, our method can adapt the task sequence, replan its original trajectory, and conduct collision avoidance for enhanced HRC. 

For the task adaptation, the task sequence for the robot to complete is a set of subtasks $\mathbf{S}$ such that $\mathbf{S} = \{s_1,s_2, \ldots, s_k \}$ where $k$ is the number of subtasks. Once a new human intention $T$ is predicted, the robot removes the $s_i$ associated with $T$ from its task sequence. This task adaptation ensures that the robot avoids redundant actions and focuses on completing the remaining tasks efficiently.

To aid in the motion planning aspect of HRC, a virtual ellipsoid is built between $T$ and the dominant wrist of the human. By building the virtual obstacle derived from the human intent prediction, the robot can follow an ordered scheme while also safely avoiding the predicted human's arm trajectory. 
The center of the virtual ellipsoid $\mathbf{c}$ can be derived using

\begin{equation}
{\mathbf{c}} = \frac{\mathbf{w} + \mathbf{T}}{2}
\label{eqn:centere}
\end{equation}

where ${\mathbf{w}}$ is the position vector of the human's wrist, and ${\mathbf{T}}$ is the position vector of the human intended object. The lengths of the semi-major axis, the semi-intermediate axis, and the semi-minor axis of the virtual ellipsoid are denoted as $a$, $b$, and $c$, respectively, which can be calculated as follows:

 
\begin{equation}
a =  \frac{{\left\|{\mathbf{w}} - {\mathbf{T}} \right\|}}{2}
\end{equation}
\begin{equation}
b = r_b \cdot a
\end{equation}
\begin{equation}
c = r_c \cdot b
\end{equation}
where $r_b$ and $r_c$ are the ratios to control the shape of the virtual ellipsoid.
We arrived at $r_b = 1/2$ and $r_c= 1/3$ by testing what works best for the real-world demonstration of a close proximity HRC task. These values can be automatically calculated based on more precise motion prediction and uncertainty estimation of human motion.

The virtual obstacle is an ellipsoid composed of sphere primitives for simplicity of real-time implementation. The ellipsoid adapts in real-time based on the human's intent. Thus, the ellipsoidal virtual obstacle is defined as

\begin{equation}
    \frac{{(x')^2}}{{a^2}} + \frac{{(y')^2}}{{b^2}} + \frac{{(z')^2}}{{c^2}} \le 1
\label{eqn:ellipsoideqn}
\end{equation} 

where $(x',y',z')$ represents the coordinates of a point in the transformed coordinate system, where the 3D ellipsoid is a standard-position ellipsoid. Our method is very robust and implements a virtual ellipse at any angle that can be observed from the camera. The constructed virtual obstacle is used with a motion planner to consider human's future motion and avoid collision with the human in a proactive manner.

\section{Experimental Setup}
To demonstrate the effectiveness of our proposed framework in Fig. \ref{fig:impl}, we developed an experimental setup for a real-world demonstration and collected a novel dataset for systematic testing.

\subsection{Hardware Setup}
The experimental hardware setup for our HRC tasks consists of several components to ensure effective interaction between the human operator and the robotic system. The workspace includes a black table where all objects used in the tasks are placed. Objects are arranged in positions that allow the UR5 robot to reach and interact with them. The table serves as the primary interaction zone. At one end of the table, a UR5 robot is equipped with a robotiq end-effector, enabling it to grasp and manipulate diverse objects. We use one Intel RealSense depth camera, the D455, as the sensor that is mounted on the frame of the setup in front of the human so that the subject's face and hands are visible along with the setup. 

A computer with two GTX Titan X GPU are connected to both the camera and the robot. The RGB-D frames from the camera are processed with MediaPipe \cite{mediapipe} and then transformed from the camera frame to the robot's base frame to extract the key points on the human's head and hand. We use MoveIt!, CHOMP motion planner, and ROS Noetic for testing and validation. The robot actions are generated and sent from the computer to the robot via an ethernet cable.

\begin{figure}[t]
\centering
\includegraphics[width= 0.95\columnwidth]
{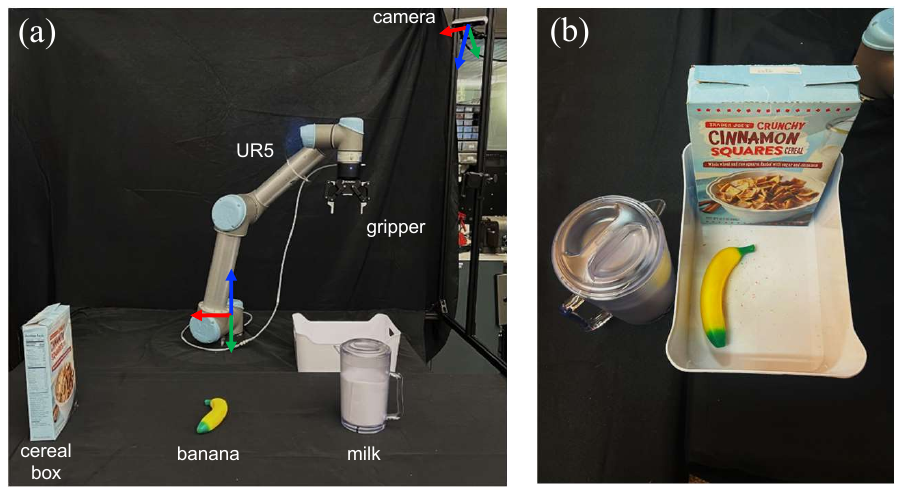}
\caption{The illustration of (a) the experimental setup and the initial state of the HRC task; (b) the goal of the HRC task. The objects of interest, potential targets, are a cereal box, a banana, and a milk jug. $x$-, $y$-, and $z$- axis are shown with red, green, and blue arrows, respectively.}
\label{fig:goalie}
\end{figure}

\subsection{Dataset Collection and Testing}
To the best of our knowledge, a dataset that provides the head key points, hand key points, and object affordance information necessary for our framework's calculations does not exist. For this reason, we created and collected our dataset called \textbf{M}IT \textbf{R}e\textbf{L}evance Dataset $(MRL)$. 
 
In the MRL dataset, the key features of the human are collected as they complete a task. The objects in the operating space are horizontally separated by 12 inches. We collected a total of 30 trajectories, 10 trajectories for each object the human is trying to grab. Our videos run at 30 FPS with a resolution of 1280$*$800 to better capture the details. The Intel D455 camera achieves 98\% accuracy depth readings within 4 meters \cite{intel}. The task in the dataset is completed less than 2 meters from the camera as shown in Fig. \ref{fig:goalie} to collect accurate depth data. The human intention of grasping and the object affordance are manually annotated. The head orientation, the hand orientation, and the hand movement are automatically annotated with our processing pipeline for each frame. 

\begin{figure}[t]
\centering
\includegraphics[width= 0.75\columnwidth]
{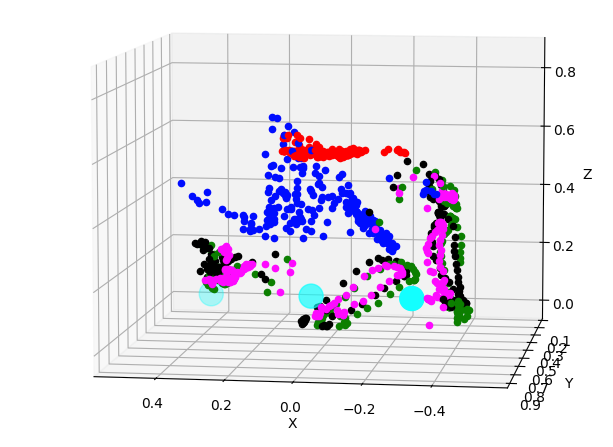}
\caption{An example of the subject's head orientation and right hand position for the task of making a bowl of cereal. The cyan spheres represent the location of the objects in the scene. The black, pink, and green points represent the three keypoints of the subject’s hand that are being tracked. The
red and blue points represent the two endpoints for the vector representing head orientation.}
\label{fig:trajref}
\end{figure}

An illustration of the key points tracked during the execution of an HRC task is shown in Fig. \ref{fig:trajref}. A total of three trajectories are shown, and our MRL dataset contains all pertinent information shown in Fig. \ref{fig:trajref} for various trajectories. 

 We used our $MRL$ dataset to learn the best CPT for HRC tasks relating to tabletop scenarios, as shown in the offline portion of Fig. \ref{fig:impl}. From our dataset, we averaged the performance of 20 observed trajectories to comprise our final CPT. The CPT provides the probability distribution of the objects for each combination of the parent nodes in Fig. \ref{fig:impl}. We used a total of 10 new trajectories, with the updated CPT, in our testing results. CPTs can also be made based on expert knowledge.

 \subsection{Demonstration Setup}
To demonstrate our methodology, we focus on the task of making a bowl of cereal. The UR5 is initially set to complete the task without human intervention. Once the human comes into the scene, BI is calculated and the robot starts adapting its behavior to proactively complete the task. The objects needed to demonstrate this scenario are a bowl, cereal box, milk, and banana. The bowl is placed next to the base of the robot and its location is also set as the drop location for the rest of the objects. We set the steps to grab milk, banana, and cereal to the bowl location in a sequence.



Fig. \ref{fig:goalie} illustrates the demonstration setup. The UR5 and human bring the necessary items to the bin based on the sequential steps related to making a bowl of cereal and the state of the human intent.

\section{Results}

To validate our results, we systematically test BI with our dataset $MRL$ and a real-world demonstration, which proves the effectiveness of our intent prediction framework and its applicability to collision avoidance and task replanning.

\subsection{Evaluation with MRL dataset}
We use the weighted average version of three metrics to evaluate our BN framework: accuracy, precision, and F1 score. The weight averaging is to offer a more balanced perspective for our multi-class intent prediction. The weight we used is a percentage that measures the occurrences of a specific class in a dataset relative to all classes available. 
We compare the proposed framework to its basic variations labeled as baselines in Table \ref{table:results}. The baseline versions only account for one modality of information: the head orientation, hand velocity, or hand orientation.

\begin{table}[h] 
\captionsetup{position=top} 
\caption{Classification results for Bayesian Intention on $MRL$ dataset. The task is to make a bowl of cereal. Using multiple modalities of information drastically improves the three metrics of intent predictions.}
\label{table:results}
\centering
\begin{tabular}{cccc} 
 \toprule
 Method & Accuracy & Precision & F1 Score \\
 \midrule
 Head Baseline & 43.45  &  18.88  & 26.32   \\ 
 Hand Orientation Baseline & 48.25  &  67.36  & 55.98   \\ 
 Hand Velocity Baseline & 31.01  &  41.31  & 33.53   \\ 
 Bayesian Intention & $\mathbf{89.55}$ &  $\mathbf{91.80}$  & $\mathbf{89.77}$ \\ 
 \bottomrule
\end{tabular}
\end{table}


As shown in Table \ref{table:results}, including multiple modalities of information greatly improves intent prediction. Our BI model achieves an F1 score of 89.77 which speaks to its well-balanced performance.  Considering the three modalities of information improves the F1 score by 60\% while the accuracy improves by 85\%  from the best baseline method. In addition, the average time for an accurate intent prediction is 2.69 ms. The baseline methods take less time to predict human intent, however the prediction is incorrect the majority of the time. Our intent prediction Bayesian model is very fast and accurate.

\begin{figure}[t]
\centering
\includegraphics[width= 0.6\columnwidth]
{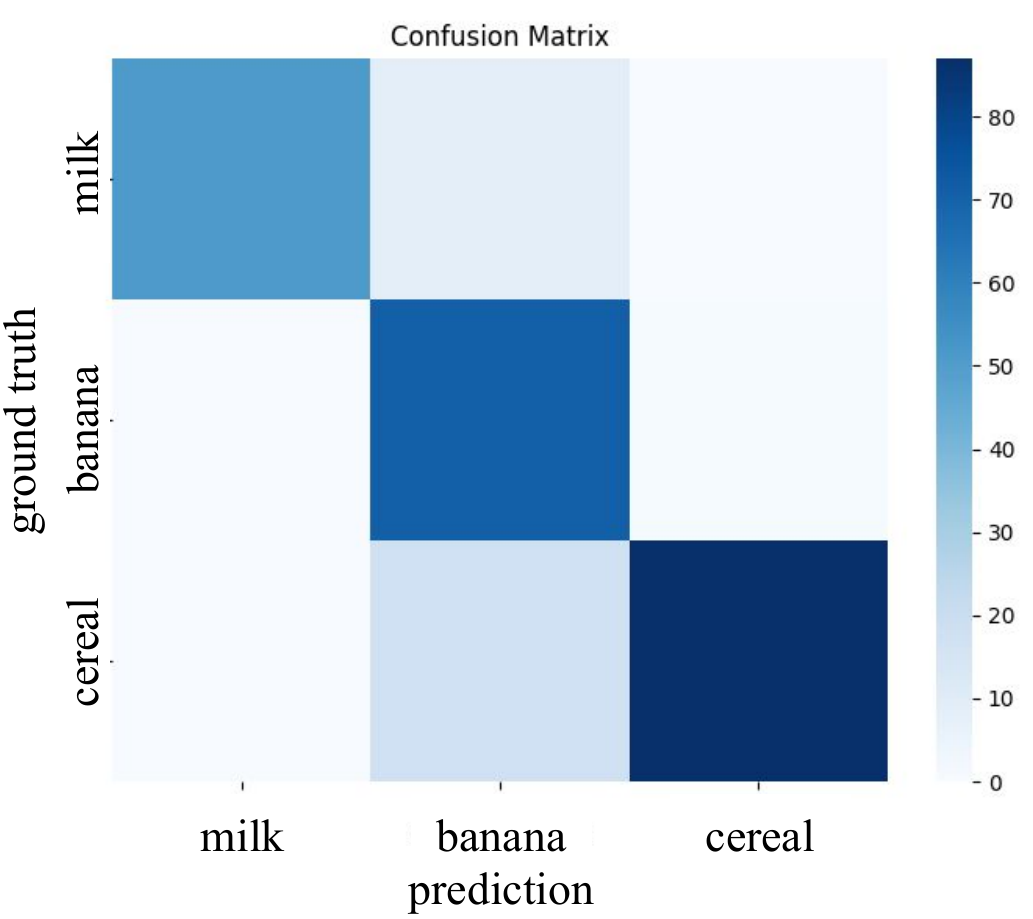}
\caption{Confusion matrix depicting the human intentions when the robot is not involved in the task. The main diagonal highlights BI's impressive ability to predict human intent accurately.}
\label{fig:confusionmfornorobot}
\end{figure}

\begin{figure*}[!t]
\centering
\includegraphics[width= 0.95\textwidth]
{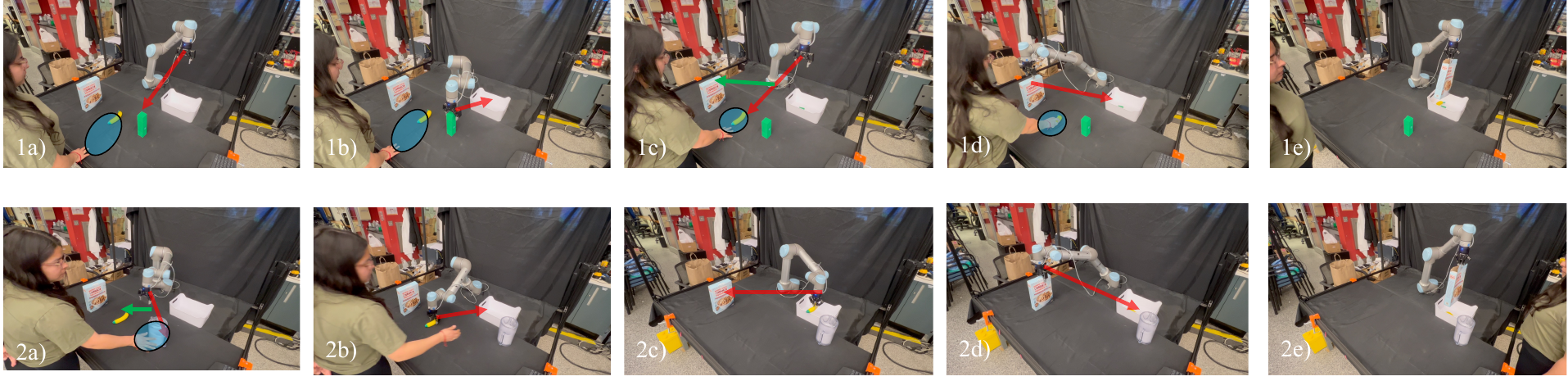}
\caption{ \small Demonstration Results. Both scenarios showcase task sequence replanning and collision avoidance. The red arrows indicate the robot's intended direction of motion while the green arrows show a redirection of the robot's path. The blue ellipses represent the virtual obstacle made to protect the human.The first row shows the case where the human grabs the banana while the robot grabs the other two items. The second row shows the case where the human grabs the milk and the robot grabs the rest of the items. The green cubes represent the milk in the first row, and they were used to facilitate an easier grasp for the robot's gripper.}
\label{fig:demo}
\end{figure*}

\begin{figure}[]
\centering
\includegraphics[width= 0.95\columnwidth]
{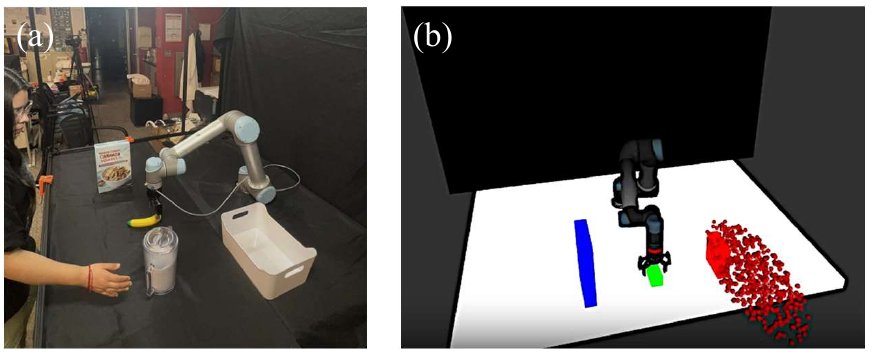}
\caption{Illustration of (a) a real-life scenario and (b) its corresponding visualization in a simulator. In the simulator, the blue, green, and red boxes represent cereal boxes, bananas, and milk jugs, respectively. The red spheres make up the virtual obstacle.}
\label{fig:scenario}
\end{figure}

Fig. \ref{fig:confusionmfornorobot} shows the intent prediction confusion matrix. The most colorful parts of the confusion matrix are along the main diagonal, which further showcases our method of predicting true positives for most cases. As shown from the top and bottom centered squares in Fig. \ref{fig:confusionmfornorobot}, our BI method predicts the human intent to be banana when the ground truth is another object in few cases. An explanation for this behavior is that the human's hand movement may contain curves that can be very influential in terms of the object that the velocity vector most closely corresponds to. For example, if the human moves their arm in clockwise direction from its start point, which is in front of the UR5, but their intent is the milk then this type of motion will cause incorrect predictions until the hand motion is more aligned to the milk. Furthermore, the affordance of the cereal box complies to either a neutral or flexion hand orientation which also contributes to the higher numbers of false positives for the banana column in Fig. \ref{fig:confusionmfornorobot}. For example, if the human directly moves on from grabbing the milk to grabbing the cereal, then that motion may be interpreted as grabbing the banana, given the direction of the velocity vector and the dual affordance that the cereal box has.

\subsection{Demonstration Evaluation}

The demonstration results are shown in Fig. \ref{fig:demo} and Fig. \ref{fig:scenario}. For this scenario, the task sequence $\mathbf{S}$ is composed of $(s_1, s_2, s_3)$ where $s_1$, $s_2$, $s_3$ represents grabbing milk, banana, and cereal box, respectively, in a sequence. Our BI model is running continuously while the task is being completed. In Fig. \ref{fig:demo}, two scenarios are depicted where the human reaches for the banana or milk and the robot successfully replans its trajectory to move onto the next subtask. This type of behavior is crucial for safe and intuitive collaboration between humans and robots. In 1a, the robot reaches for the milk. In 1b, the robot is transferring the milk to the drop location. In 1c, the robot recognizes a future collision if it continues to grab the banana as planned, so it moves on to grabbing the cereal box. In 1d, the robot plans its trajectory to complete the task. In 1e, the human is ready to start making a bowl of cereal with all the necessary items in one place. In 2a, the robot is originally reaching for the milk but replans to follow the green arrow to avoid collision with the human and advances to the next subtask. In 2b-2d, the robot is shown completing the task with the human. In 2e, they are ready to make a bowl of cereal with all the necessary items in one place.




In Fig. \ref{fig:scenario}, an illustration of the virtual obstacle can be seen as the UR5 replans its task and motion to grabbing the banana instead of the milk.  The hand's orientation in Fig. \ref{fig:scenario} exemplifies how object affordance influences hand orientation, which verifies our hypothesis. One does not expect to grab a milk jug with a hovering orientation as the base is too large to grasp in a top down manner.

The demonstration results show our method can smoothly avoid the contour of the virtual obstacle, making it effective in dynamic motion planning and task replanning in real-time.

\section{Conclusion}
We introduce the BI framework, for modeling the interplay between objects, task, and the human model, to predict human intent. Our results prove BI's significant potential, across various HRC domains, to infer human intent using three modalities of information. This differs from previous statistical modeling works that only use gaze and skeletal position data to predict intent. Our BN is built with clear causality between variables that help explain the logic behind human intent in general HRC scenarios. We assess our BI framework on our new $MRL$ dataset that tracks the head and hand key points for human intent prediction. Our experimental results, demonstrated through an HRC task with a UR5 robot, show that the BI framework significantly enhances real-time human intent prediction and collision avoidance, achieving a 36\% increase in precision, a 60\% increase in F1 Score, and an 85\% increase in accuracy over the best baseline method! Our framework's fast and accurate intent predictions enhance the robot's response by improving safety and efficiency in HRC environments.


\addtolength{\textheight}{-13cm}   




\newpage

\bibliographystyle{IEEEtran}
\bibliography{IEEEabrv,references}

\end{document}